\pdfoutput=1

\documentclass[11pt]{article}

\usepackage[final]{acl}

\usepackage{times}
\usepackage{latexsym}

\usepackage[T1]{fontenc}

\usepackage[utf8]{inputenc}

\usepackage{microtype}

\usepackage{subcaption}
\usepackage{amsfonts}
\usepackage{amsmath}
\usepackage{multirow}
\usepackage{multicol}
\usepackage{makecell}
\usepackage{adjustbox}
\usepackage{tabularx}

\usepackage{inconsolata}
\usepackage{graphicx}

\title{SS-MPC: A Sequence-Structured Multi-Party Conversation System}

\author{Yoonjin Jang \\
  SungKyunKwan University \\
  Republic of Korea \\
  \texttt{yoonjinjang98@gmail.com} \\\And
  Keunha Kim \\
  SungKyunKwan University \\
  Republic of Korea \\
  \texttt{keunhakim98@gmail.com} \\\And
  Youngjoong Ko\thanks{Corresponding author} \\
  SungKyunKwan University \\
  Republic of Korea \\
  \texttt{yjko@skku.edu} \\}

\begin{document}
\maketitle
\begin{abstract}
Recent Multi-Party Conversation (MPC) models typically rely on graph-based approaches to capture dialogue structures. However, these methods have limitations, such as information loss during the projection of utterances into structural embeddings and constraints in leveraging pre-trained language models directly. In this paper, we propose \textbf{SS-MPC}, a response generation model for MPC that eliminates the need for explicit graph structures. Unlike existing models that depend on graphs to analyze conversation structures, SS-MPC internally encodes the dialogue structure as a sequential input, enabling direct utilization of pre-trained language models. Experimental results show that \textbf{SS-MPC} achieves \textbf{15.60\% BLEU-1} and \textbf{12.44\% ROUGE-L} score, outperforming the current state-of-the-art MPC response generation model by \textbf{3.91\%p} in \textbf{BLEU-1} and \textbf{0.62\%p} in \textbf{ROUGE-L}. Additionally, human evaluation confirms that SS-MPC generates more fluent and accurate responses compared to existing MPC models.
\end{abstract}

\section{Introduction}
The rapid development of the Internet and the social media platforms have changed the way people communicate with each other and created new forms of interaction. In particular, Multi-Party Conversation (MPC), conversation in which multiple people freely exchanges opinions at the same time, is increasingly becoming common. MPC occurs on many platforms, such as group chats, online forums, and comment sections on social media. Recent research trends show that analysis and response generation on MPC is in its infancy, and the importance and need for them is increasingly being emphasized \cite{Park2020Investigating, Anjum2020VertextAE}.

\begin{figure}[t!]
    \centering
    \includegraphics[width=\linewidth]{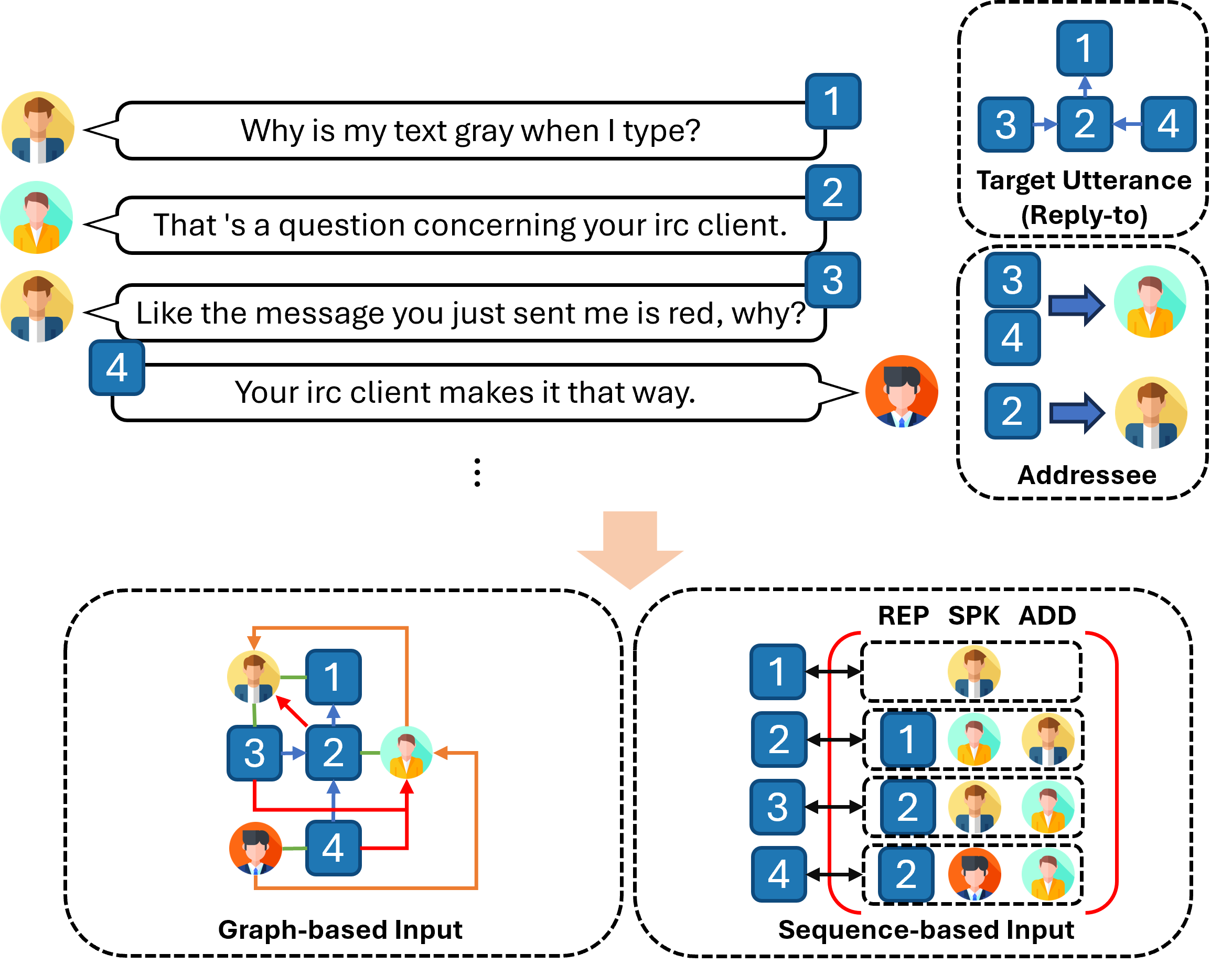}
    \caption{An example of data in MPC Dataset (Ubuntu IRC Benchmark Dataset). The dataset is constructed of context and structural information. Context consists of utterances, and structural information consists of speaker information, target-utterance relation and addressee relation of each utterance.}
    \label{fig:dataset}
\end{figure}

MPC have flexibility to allow multiple speakers to participate in a conversation in no particular order or rule. These attributes of MPC complicate the flow and structure of the conversation compared to one-on-one conversations, and create additional challenges in understanding the context and intent of the utterances and speakers. Each speaker brings a unique context and intent to the conversation, and he or she must understand and process the context in which a particular utterance is being delivered to whom. 
However, because the addressee of an utterance is often unclear in MPC, predicting or analyzing the structure of the dialogue is one of the major challenges for MPC. In addition, because multiple participants are simultaneously expressing their opinions and interacting, the topic and flow of the conversation are likely to change frequently. Unlike one-on-one conversations, these characteristics create the additional challenge of closely tracking and managing the context of each utterance in MPC. For these reasons, developing systems for generating responses to MPC is considered one of the most challenging areas of current dialogue system research.

To address this complexity, MPC datasets typically include more structural conversation information for each utterance \cite{lowe-etal-2015-ubuntu}.
For example, as shown in Figure \ref{fig:dataset}, each utterance contains the structural information such as speaker information, which tells us who is speaking, the addressee information, which tells us to whom the utterance is addressed, and the target-utterance information, which indicates which utterance the current utterance is responding to.
The target-utterance relationship is typically linked to only one previous utterance in the conversation history, and the addressee is semantically the same as the speaker of the target-utterance. This structural information is critical for dialogue systems to understand and learn about the complexity of MPC, and it helps MPC response generation models generate responses that are appropriate for a given context.

However, traditional language models using sequential input have found that it is very difficult to express the structure of these conversations. Recent work \cite{gu-etal-2022-hetermpc,gu-etal-2023-madnet} has attempted to solve this problem using graphs. By representing utterances, speakers, and the relationships between utterances and speakers as graphs, and interpreting them through a graph encoder, it was possible to train a model to recognize the structure of a conversation and generate responses accordingly. But this still limits the use of pre-trained language models itself, since there are no such pre-trained graph-encoder models tuned properly to analyze the structure of conversations. If we partially employ randomly initialized graph-encoder, it may break the embedding space of the entire pre-trained model.

In this paper, we introduce \textbf{S}equence-\textbf{S}tructured \textbf{MPC} (SS-MPC), a response generation system for multi-party conversations that leverages the encoder-decoder architecture of the transformer while replacing the role of a graph encoder with a soft prompt. Instead of explicitly encoding conversational structures using a graph encoder, SS-MPC effectively integrates structural information through well-designed soft prompts within the standard encoder-decoder framework. This approach eliminates the need for additional model components, accelerates the training process, and achieves superior performance compared to existing models.

Futhermore, unlike additinal models that require MPC analysis for the response generation, SS-MPC has the advantage of being able to analyze conversations and generate responses at once by end-to-end, enabling immediate usage in real-world MPC environments. By learning the conversation structure such as the correct target-utterance and addressee information for each utterance during the training process, the model can make appropriate inferences and generate the final response even when some of the correct target-utterance and addressee information is omitted during the actual inference process. This can be done because the SS-MPC contains the post-training process with the way masking the information partially, which means that the model itself is already trained for the ability to predict the omitted target-utterance or addressee information.

Our contributions are summarized as follows:
\begin{itemize}
    \item We propose a novel method to train language models for MPC response generation without graph structures, which leverages sequence-structured inputs to internally represent the interaction flow in the dialogue.
    \item The proposed model can be used in real-world MPC environments easily because the model can simultaneously analyze conversations and generate responses using an end-to-end framework.
    \item Experimental results show that the proposed model performs better than the previous SOTA model. In addition, our various analysis results provide directions for future research in multi-party dialogues.
\end{itemize}

\section{Related Work} 

\subsection{Multi-Party Dialogue Structural Analysis}
The task of predicting relationships between speakers to analyze the structure of MPC began in 2016. \citet{ouchi-tsuboi-2016-addressee} first proposed the addressee prediction and utterance selection task. To study this task, they first hand-created a dataset of MPC using log transcripts from Ubuntu IRC channels, and then utilized RNNs to perform the proposed task. Later, \citet{Zhang2018Addressee} proposed SI-RNN, which updates speaker embeddings based on roles for addressee prediction. \citet{Meng2017TowardsNS} also proposed a speaker classification task to model the relationships between speakers. Meanwhile, for the MPC response selection task, \citet{wang-etal-2020-response} proposed to track dynamic topics, and then a who-to-whom (W2W) model \cite{le-etal-2019-speaking} was proposed to predict the addressees of all utterances in a conversation. \citet{Gu2021MPCBERTAP} proposed the MPC-BERT model, which utilizes multiple MPC learning methods to learn the complex interactions between recent utterances and interlocutors, and it performs post-training for MPC tasks. In addition, \citet{Gu2023GIFTGF} proposed the GIFT model to help fine-tuning for MPC tasks with only simple scalar parameters on the attentions.

However, all the methodologies proposed in the above works have the limitation that they utilize the utterances to predict the addressee information of each utterance. This makes it difficult to utilize these methodologies for response generation models in real-world MPC environments.

\subsection{Multi-Party Dialogue Response Generation}
Along with these MPC tasks, there has been a parallel research on MPC response generation, which is the task of generating responses to a multi-party dialog. \citet{Hu2019GSNAG} proposed a graph structure network (GSN) to model the graphical information flow for response generation. Later, Heter-MPC \cite{gu-etal-2022-hetermpc} was proposed to model complex interactions between utterances and interlocutors as graphs. This paper used graphs with two types of nodes and six types of edges to model the structure of multi-party conversations. \citet{li-zhao-2023-em} utilized the Expectation-Maximization (EM) algorithm in pre-training to predict the missing addressee information in the dataset. However, they still suffer from the drawback that the fine-tuning process only allows for an ideal setup where all addressees are labeled. To overcome this, MADnet \cite{gu-etal-2023-madnet} utilizes the EM algorithm in the model of \citet{gu-etal-2022-hetermpc} to directly predict and supplement the missing addressee information for training and response generation.

\begin{figure*}[t!]
    \centering
    \includegraphics[width=\textwidth]{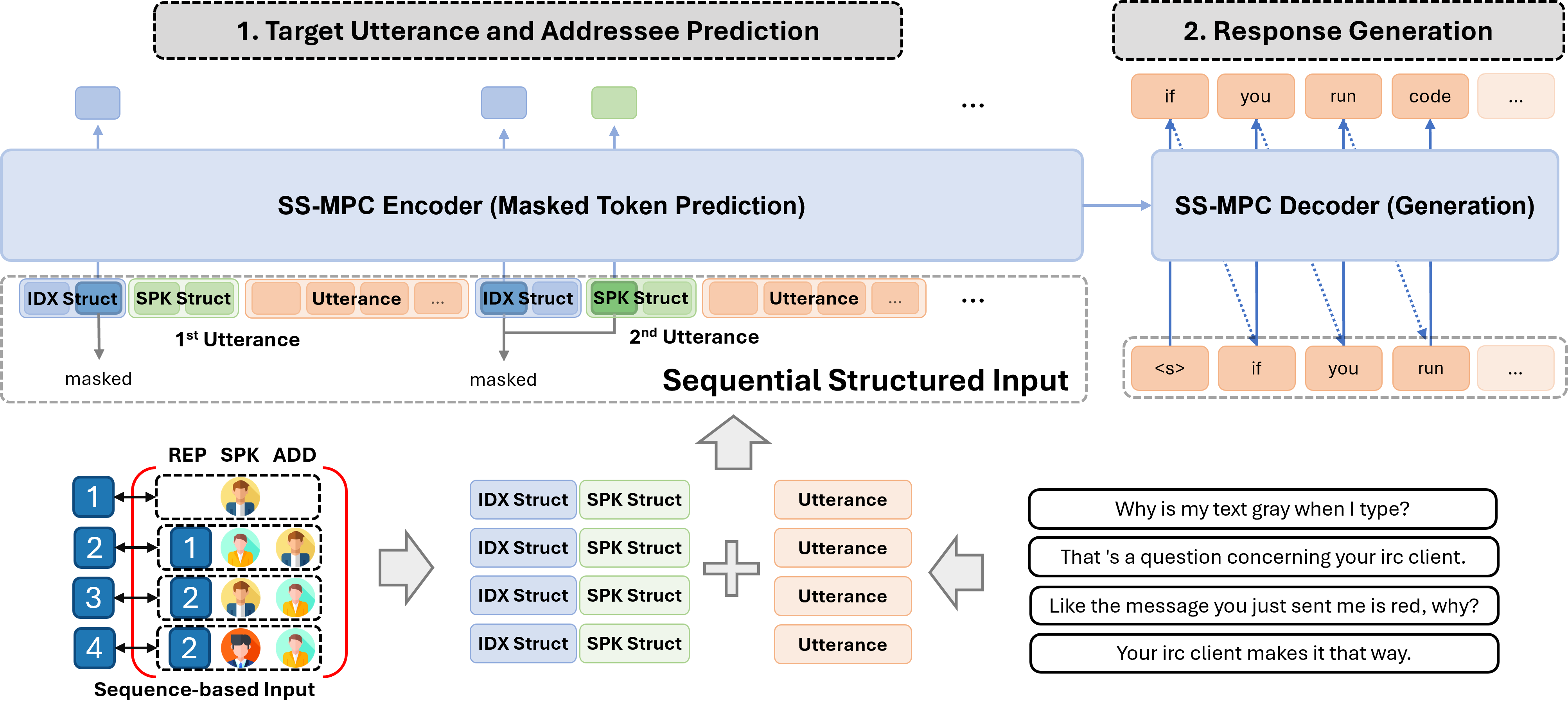}
    \caption{The overview of the SS-MPC. The encoder part is expected to analyze the dialogue and predict the structural information in dialogue. The decoder part is expected to generate the final response with using the information analyzed in encoder.}
    \label{fig:overview}
\end{figure*}

\section{Methodology}
In this paper, we propose SS-MPC, a novel MPC response generation model with the encoder-decoder structure of transformer. Here, we describe its input, output, and the training process.

\subsection{Preliminaries}

In a typical response generation task, the goal is to generate a final response \(\Bar{r}\) for a given conversation history \(h\). In addition, the traditional MPC response generation task utilizes not only the dialogue history but also the dialogue structural information \(C\). \(C\) consists of the speaker \(c_i\), addressee \(a_i\), and target-utterance \(u_i\) for each of the utterances.

\begin{equation}
C=\{c_1, c_2, \ldots, c_n\}
\end{equation}
\begin{equation}
c_i = \{s_i, a_i, u_i\}
\end{equation}
where the number of the utterances is $n$ and $1 \leq i \leq n$.

In general, the MPC datasets provide \(C\), and the MPC models perform the task of finding the most appropriate response \(\Bar{r}\) based on the given \(h\) and \(C\) information. The response tokens are generated in an auto-regressive way. This can be formulated as follows:
\begin{multline}
\Bar{r}=\underset{r}{argmax} \ P(r|h,C;\theta)\\
= \underset{r}{argmax} \underset{t=1}{\overset{|r|}{\prod}} \ P(r_t|h,C, r_{<t};\theta)
\end{multline}
where \(r_t\) means the $t$-th token, and \(r_{<t}\) means the previous tokens of the final response \(r\). 

The existing MPC models have utilized graph-based models to use \(C\). In this process, there can be a loss of information when aligning the embedding space of \(h\) with \(C\) and using a graph encoder after random initialization.
\begin{equation}
\Bar{r}=\underset{r}{argmax} \ P_{Dec}(r|GraphEnc(h,C);\theta)\\
\end{equation}
where \(P_{Dec}\) means the probability of each token computed by the decoder of the model, and \(GraphEnc(\cdot)\) means the embedding created
by the graph encoder of the model. 

In the case of SS-MPC, we use the sequence-structured input of each utterance, so we can utilize the full information of dialogue history and dialogue structural information without using a graph encoder. 
\begin{equation}
\Bar{r}=\underset{r}{argmax} \ P_{Dec}(r|Enc(h,s(C));\theta)\\
\end{equation}
where \(s(\cdot)\) is the dialogue structuralization, which means transforming original input as the sequence-structured input containing the dialogue structure information internally.

Furthermore, we should consider the situation where the lack of structural information exists in MPC. Graph-encoder needs a fully connected graph since the lack of edges means that unconnected nodes can confuse the model in encoding. But SS-MPC does not need any change in model structure or additional training in this situation because the model is already trained with the way masking the information partially. The inference process of the SS-MPC is as follows:

\begin{equation}
\Bar{r},\Bar{C}= \\
\underset{r}{argmax} \ P(r|h,s(C_{omit});\theta)
\label{eq:omitted}
\end{equation}
where \(C_{omit}\) is the partially omitted structural information in MPC, and \(\Bar{C}\) means the predicted structural information for \(C_{omit}\).

\subsection{Overview of SS-MPC}
Figure \ref{fig:overview} shows the overview of the SS-MPC. It utilizes the encoder-decoder structure of the transformer because the transformer encoder is commonly used to analyze the structure of conversations \cite{Shen2020DialogXLAX,mehri-etal-2019-pretraining}, and we want to leverage the strengths of these encoders to analyze the structure of conversations while allowing the decoder to focus on generating the actual responses.

The first step to utilize sequence-structured input instead of graph-structured input is to insert the conversation structure into a sequential form. To do this, we add three soft prompt tokens to the model tokenizer as MPC structure tokens to represent the structure of the conversation by adding information about each utterance. The embeddings of new tokens act as soft prompts for each utterance, which reflect the structural information of the conversation.

Then we post-train only the encoder with the task of predicting dialogue structure to improve the encoder's ability to interpret the meaning of the added structure tokens and analyze the context with the dialogue structure. Through this process, the model not only learns the meaning of the added MPC structure tokens, but also learns to predict the partially omitted dialogue structure information of each utterance using the information of previous utterances and speakers.

\subsection{MPC Structure tokens}
To represent the speaker, addressee, and target-utterance for each utterance as a dialogue structure, three kinds of soft prompt tokens are added to the model tokenizer, called by MPC structure tokens; their embeddings are randomly initialized before training. The added MPC structure tokens are as follows:
\vspace{-0.3em}
\begin{itemize}
    \setlength\itemsep{-0.4em}
    \item Index structure token
    \item Speaker structure token
    \item Structure masking token
\end{itemize}
\paragraph{Index structure token} These tokens are developed to distinguish the order of the conversation; they are designed by indicating the order of the each utterance by number, such as "$[IDX_1]$", "\([IDX_2]\)", \ldots ,"\([IDX_n]\)". The target-utterance can be also represented by specifying the index of the target-utterance.

\paragraph{Speaker structure token}
These tokens are to distinguish speakers in a conversation based on the order in which they appear; they are designed to distinguish between speakers, such as "\([SPK_1]\)", "\([SPK_2]\)", \ldots , "\([SPK_m]\)". The addressee information for each utterance can be expressed by specifying the speaker information via the corresponding token. 

\paragraph{Structure masking token}
"$[Mask_{\text{IDX}}]$" and "$[Mask_{\text{SPK}}]$" tokens are added to mask MPC structure information. The "$[Mask_{\text{IDX}}]$" token masks the index and the target-utterance information, and the "$[Mask_{\text{SPK}}]$" token masks the speaker and the addressee information in a given utterance.

\begin{figure}[t!]
    \centering
    \includegraphics[width=\linewidth]{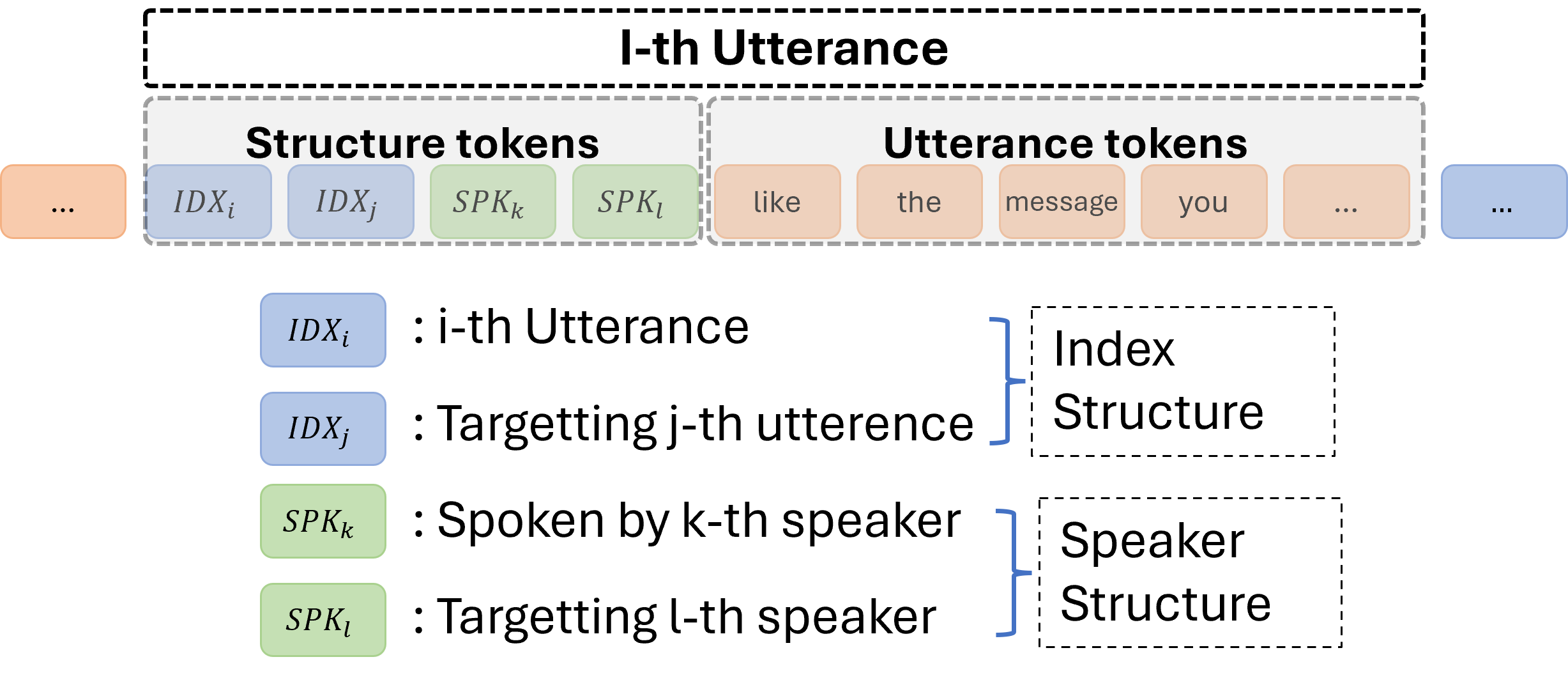}
    \caption{An example of the sequence-structure template for an utterance. Two index structure tokens which represents the utterance's index and the target-utterance's index, and two speaker structure tokens which represents speaker and addressee of the utterance are added as prefix to the tokenized utterance tokens.}
    \label{fig:structuring}
\end{figure}

\vspace{1em}
Since there is not a target-utterance and addressee for the first utterance, we add "$[IDX_{\text{None}}]$" token and "\([SPK_{\text{None}}]\)" token to indicate that it has no target-utterance or addressee information.
 
\subsection{Dialogue Structuralization}
Dialogues are sequence-structured using MPC structure tokens that are added to express the structure of the conversation in a sequential form. The entire conversation can be broken down into utterances, and each of the utterances consists of the structure tokens and utterance tokens. Note that, for the final response, the utterance tokens are omitted because it is the answer which the model should generate.

In Figure \ref{fig:structuring}, an example is shown for the $i$-th utterance of the entire conversation. The token that indicates that the $i$-th utterance is labeled by "\([IDX_i]\)". This $i$-th utterance is responding to the $j$-th utterance, which labeled by the second token "\([IDX_j]\)". Furthermore, the $i$-th utterance is being uttered by the speaker-$k$ and is being answered to the speaker-$l$, which is represented by the tokens "\([SPK_k]\)" and "\([SPK_l]\)" in the following sequence. Thus we can set the sequence-structure template of $i$-th utterance as follows: 
\begin{multline}
S_i = (\overbrace{[IDX_i];[IDX_j];[SPK_k];[SPK_l]}^{structure\ tokens};\\\overbrace{[token_1];[token_2]...}^{utterance\ tokens})
\end{multline}
where \(1 \le i \le n\) for dialogue with $n$ utterances.

Then the model should generate the final response, so its structure information is inputted. The sequence-structure template is set as follows:
\begin{equation}
S_{r} = (\overbrace{[IDX_{r}];[IDX_{t}];[SPK_{s}];[SPK_{a}]}^{response\ structure\ tokens})
\label{eq:final_response}
\end{equation}
where "$[IDX_{r}]$","$[IDX_{t}]$","$[SPK_{s}]$", and "$[SPK_{a}]$" tokens means the index, target-utterance index, speaker, and addressee of the current response utterance.

The sequence-structure templates of whole utterances in a dialogue are concatenated as a sequence-structured input \(S = (S_1;S_2;...;S_{n-1};S_n;S_{r})\). In addition, the structural information of the input can be masked using the "$[Mask_{\text{IDX}}]$" and "$[Mask_{\text{SPK}}]$" tokens. This masking approach is performed when target-utterance and addressee information has to be predicted in the conversation. In particular, the encoder is post-trained using the masking approach. 

\subsection{Post-training for Encoder}
SS-MPC with the addition of soft prompt tokens needs to carry out post-training to obtain better embeddings of soft prompt tokens containing semantic and contextual information from context tokens. 
In particular, masking is performed with a probability of hyper-parameter $p$\% for the structure tokens of each utterances, and the encoder predicts the correct answer for the masked structure tokens. To predict the structure token, the LM head of post-training for encoder shares the parameter with the LM head in decoder since they generate the same type of tokens; For the utterance tokens, we train the encoder to generate the tokens itself. The formulation of the loss function for post-training is as follows:

\begin{equation}
\mathcal{L}_{post} = -\sum_{i} \log P(x_i | X_{masked}; \phi)
\end{equation}

\noindent where \(i\) means each position of the input, \(x_i\) denotes the original encoder input token of the \(i\)-th position, for the masked encoder input $X_{masked}$. And the $\phi$ is the parameter of the SS-MPC encoder.

\subsection{Fine-Tuning Model}
SS-MPC is fine-tuned after post-training to perform the task of generating the final response. Fine-tuning is same as the learning process of a typical transformer encoder-decoder model. The difference is that the SS-MPC utilizes sequence-structured input.

\begin{equation}
\mathcal{L}=-\overset{n}{\underset{i=1}{\sum}}logP(r_i|r_{<i},X;\theta)
\end{equation}

\noindent where $r_i$ is the $i$-th final response token, $X$ is the sequence-structured encoder input, and $\theta$ is the parameter of the SS-MPC. 

\section{Experiments}
\paragraph{Dataset}
To evaluate the performance of the proposed SS-MPC model, we utilize the Ubuntu IRC benchmark dataset, which has originally released by \citet{ouchi-tsuboi-2016-addressee} and \citet{Hu2019GSNAG}, and has been widely using for various MPC tasks. This dataset comprises user conversations from the Internet Relay Chat (IRC) channel of the Ubuntu homepage. 

Two Ubuntu IRC Benchmark datasets are used in the experiments as follows:

\textbf{Ubuntu IRC (2016):} The dataset released by \citet{ouchi-tsuboi-2016-addressee}\footnote{We adopt the refined version provided by \citet{le-etal-2019-speaking}, which is released on \url{https://github.com/lxchtan/HeterMPC} \cite{gu-etal-2022-hetermpc}} has some missing structural information in the dataset. We construct the sequence-structured input with masking those missing information. This dataset is categorized into three subsets based on session length (Len-5, Len-10, and Len-15). We employ the Len-5 subset, following the settings of preivious studies.

\textbf{Ubuntu IRC (2019):} The dataset released by \citet{Hu2019GSNAG}\footnote{We adopt the re-emplemnted processed version of \cite{gu-etal-2022-hetermpc}, which is released on \url{https://github.com/lxchtan/HeterMPC}} includes all structure information for every utterances. The dataset is used for post-training SS-MPC both in Ubuntu IRC (2016) and Ubuntu IRC (2019). Further details on the datasets can be found in the Appendix \ref{sec:data_stats}.

\paragraph{Evaluation Metrics}
To evaluate SS-MPC, we just follow previous research and measure its performance using BLEU-1 through BLEU-4, METEOR, and ROUGE-L score for the final response. All metrics are computed using the Hugging Face \texttt{evaluate} library\footnote{\url{https://huggingface.co/docs/evaluate/index}} \cite{wolf-etal-2020-transformers}. 

\begin{table*}[t]
\centering
\resizebox{\textwidth}{!}{%
\begin{tabular}{llcccccc}
\hline
Dataset &\textbf{Models} & \textbf{BLEU-1} & \textbf{BLEU-2} & \textbf{BLEU-3} & \textbf{BLEU-4} & \textbf{METEOR} & \textbf{ROUGE-L} \\
\hline
\hline
\multirow{5}{*}{Ubuntu IRC (2016)} & GSN$\dagger$ \cite{Hu2019GSNAG}             & 10.23 & 3.57 & 1.70 & 0.97 & 4.10 & 9.91 \\
&GPT-2 \cite{radford2019language}        &8.86 & 2.69 & 1.11 & 0.61 & 7.40 & 8.53 \\
&BART \cite{lewis-etal-2020-bart}             &11.76 & 4.86 & 2.97 & 2.21 & 8.91 & 9.86\\
&MADNet   \cite{gu-etal-2023-madnet}     &  11.82 & 4.58 & 2.65 & 1.91 & 9.78 & 10.61 \\
&\textbf{SS-MPC} & \textbf{13.40} & \textbf{5.87} & \textbf{3.60} & \textbf{2.65} & \textbf{10.15} & \textbf{11.14} \\
\hline
\multirow{6}{*}{Ubuntu IRC (2019)} &GSN$\dagger$ \cite{Hu2019GSNAG}  & 6.32 & 2.28 & 1.10 & 0.61 & 3.27 & 7.39 \\
&GPT-2 \cite{radford2019language}              & 10.85 & 3.76 & 1.61 & 0.84 & 9.00 & 7.24 \\
&BART \cite{lewis-etal-2020-bart}              & 12.71 & 4.52 & 2.13 & 1.25 & 8.75 & 10.01 \\
&HeterMPC \cite{gu-etal-2022-hetermpc}         & 10.29 & 3.68 & 1.71 & 0.96 & 8.79 & 11.22 \\
&MADNet   \cite{gu-etal-2023-madnet}           & 11.69 & 4.57 & 2.33 & 1.45 & 9.48	& 11.82 \\
&\textbf{SS-MPC (Ours)}   & \textbf{15.60} & \textbf{6.62} & \textbf{3.67} & \textbf{2.44} & \textbf{10.92} & \textbf{12.44} \\
\hline
\end{tabular}
}
\caption{Performance comparison of different models on two Ubuntu IRC Benchmark datasets \cite{ouchi-tsuboi-2016-addressee,Hu2019GSNAG} with various metrics. The performance result of GSN$\dagger$ is cited from \citet{gu-etal-2023-madnet}.}
\label{tab:ubuntu2016&2019}
\end{table*}

\begin{table}[h!]
\centering
\resizebox{\linewidth}{!}{%
\begin{tabular}{lccccc}
\hline
\textbf{Human Evaluation} &&&& \textbf{Score} & \textbf{Kappa} \\
\hline
Gold Label &&&& 1.91 & 0.51 \\
\hline
GPT-2 \cite{Hu2019GSNAG}   &&&& 0.6 & 0.50\\
BART \cite{lewis-etal-2020-bart}   &&&& 1.50 & 0.48\\
MADNet \cite{gu-etal-2023-madnet} &&&& 1.57 & 0.46 \\
\textbf{SS-MPC (Ours)} &&&& \textbf{1.84} & 0.55\\
\hline
\end{tabular}
}
\caption{Human Evaluation results on Ubuntu IRC Benchmark test set of Ubuntu IRC (2019) \citet{Hu2019GSNAG} with several MPC response generation models.}
\label{tab:human_evaluation}
\vskip -.16in
\end{table}

\paragraph{Baselines}
For the backbone model, we use \textbf{BART} \cite{lewis-etal-2020-bart} as a widely recognized transformer-based encoder-decoder model. BART leverages the encoder-decoder architecture that are well-suited for response generation as well as other generative tasks such as summarization and machine translation.

We compare our approach against the following models: \textbf{(1) GSN} \cite{Hu2019GSNAG}. GSN's core architecture consists of an utterance-level graph-structured encoder. \textbf{(2) GPT-2} \cite{radford2019language}, a unidirectional pre-trained language model. Following its original setup, all context utterances and response are concatenated by using a special token "$[SEP]$" as input.
We also compare ConvMPC with the \textbf{(3) HeterMPC} \cite{gu-etal-2022-hetermpc} and \textbf{(4) MADnet} \cite{gu-etal-2023-madnet}, which are known as SOTA models among the current MPC response generation models. The HeterMPC model structures the speaker, target utterance, and addressee relationships in the form of a heterogeneous graph to model complex MPC. To analyze the structured data, it utilizes a heterogeneous graph encoder structure that utilizes Graph Attention (GAT) operations. In the case of the MADnet model, it is a model that slightly modifies the graph input of the existing HeterMPC model and adds the EM-algorithm methodology to generate a response by inferring the missing data from the existing data through the EM-algorithm.

We further evaluate our approach on decoder based Large Language Model (LLM) in appendix \ref{sec:llm}.

\paragraph{Implementation Details}
Model parameters are initialized with \textbf{BART-large} \cite{lewis-etal-2020-bart}, which were implemented in Hugging Face's \texttt{Transformers} library \footnote{\url{https://huggingface.co/facebook/bart-large}} \cite{wolf-etal-2020-transformers}. We use AdamW \cite{Loshchilov2017DecoupledWD} for optimization, with an initial learning rate of $1e\text{-}5$ that decayed linearly. The model is trained on the two Ubuntu IRC benchmark training sets, with a maximum of 10 epochs for post-training and fine-tuning individually. We use a batch size of 8 with 2 gradient accumulation steps and select the best model based on validation performance for testing. The maximum length of the generated output is set to 50 tokens just following previous studies.

\subsection{Main Results}
Table \ref{tab:ubuntu2016&2019} shows model performances on two Ubuntu IRC Benchmark datasets, \textbf{Ubuntu IRC (2016)} \cite{ouchi-tsuboi-2016-addressee} and \textbf{Ubuntu IRC (2019)} \cite{Hu2019GSNAG}. SS-MPC achieves a significant performance improvement on both datasets compared to previous models. On the \textbf{Ubuntu IRC (2016)} dataset, SS-MPC outperforms the previous state-of-the-art (SOTA) model, MADNet, by 1.58\%p in BLEU-1, 1.29\%p in BLEU-2, 0.95\%p in BLEU-3, 0.74\%p in BLEU-4, 0.37\%p in METEOR, and 0.53\%p in ROUGE-L.

Similarly, on the \textbf{Ubuntu IRC (2019)} dataset, SS-MPC also surpasses MADNet by 3.91\%p in BLEU-1, 2.05\%p in BLEU-2, 1.34\%p in BLEU-3, 0.99\%p in BLEU-4, 1.44\%p in METEOR, and 0.62\%p in ROUGE-L. You can see the response examples generated by each model in Appendix \ref{sec:case_study}. 

\subsection{Human Evaluation}
Since quantitative metrics alone may not fully capture the quality of generated responses, we also conduct a human evaluation. Specifically, we randomly sampled 100 conversations from the Ubuntu IRC benchmark dataset and asked three graduate students to evaluate the quality of the generated responses. The evaluation focuses on three independent aspects: (1) relevance, (2) fluency, and (3) informativeness. Each judge assigned binary scores for each aspect, with the final score ranging from 0 to 3. Table \ref{tab:human_evaluation} presents the average final score of human evaluation results comparing GPT-2, BART, MADNet, and SS-MPC against the ground truth (Gold Label). In addition, Fleiss’s Kappa \cite{fleisskappa1971} is calculated to measure inter-annotator agreement. The result indicates that SS-MPC produces more relevant, fluent, and informative responses than any other model.

\begin{table}[t!]
\centering
\resizebox{\linewidth}{!}{%
\begin{tabular}{lcccccccccc}
\hline
\textbf{Masking Rates} & \textbf{BLEU-1} & \textbf{METEOR} & \textbf{ROUGE-L} \\
\hline
\hline
w/o post-training                                  & 14.22 & 9.87 & 11.84 \\
\textbf{p=25\%  (SS-MPC)}                       & \textbf{15.60} & \textbf{10.92} & \textbf{12.44} \\
p=50\%                                  & 14.30 & 9.87 & 12.04 \\
p=75\%                                  & 13.54 & 9.20 & 11.15 \\
p=100\%                                 & 13.81 & 9.45 & 11.24 \\
\hline
\end{tabular}
}
\caption{Ablation Study of the SS-MPC based on masking rate during post-training.}
\label{tab:masking}
\end{table}

\begin{table}[t!]
\centering
\resizebox{\linewidth}{!}{%
\begin{tabular}{l c c}
\hline
\textbf{Model} & \textbf{Target Utt.} & \textbf{Adr.}\\
\hline
\hline
BERT \cite{devlin-etal-2019-bert} & - & 82.88\% \\
SA-BERT \cite{sun-etal-2019-utilizing} & - & 86.98\% \\
MPC-BERT \cite{Gu2021MPCBERTAP} & - & 89.54\% \\
GIFT \cite{Gu2023GIFTGF} & - & 90.18\% \\
\textbf{SS-MPC$_{Encoder}$}  & 76.38\% & 89.92\% \\
\hline
\end{tabular}
}
\caption{Performance (precision@1) of predicting target-utterance and addressee on the test set of Ubuntu IRC (2019) \cite{Hu2019GSNAG}.}
\label{tab:accuracy}
\vskip -.16in
\end{table}

\begin{table*}[t]
\centering
\resizebox{\textwidth}{!}{%
\begin{tabular}{llcccccc}
\hline
\textbf{Models}  & \textbf{Structural Info.} &\textbf{BLEU-1} &\textbf{BLEU-2}&\textbf{BLEU-3}& \textbf{BLEU-4} & \textbf{METEOR} & \textbf{ROUGE-L} \\
\hline
\hline
MADNet   \cite{gu-etal-2023-madnet}         & \multirow{2}{*}{fully given}  & 11.69 & 4.57 & 2.33 & 1.45 & 9.48	& 11.82 \\
SS-MPC  && 15.60 & 6.62 & 3.67 & 2.44 & 10.92 & 12.44 \\
\hline
SS-MPC$_{real\text{-}world}$  & -last utt. & 13.59 & 5.47 & 3.10 & 2.14 & 9.15 & 11.03 \\
\hline
\end{tabular}
}
\caption{Performance of SS-MPC without structural information of last utterance.}
\label{tab:structure_masked}
\end{table*}

\begin{table}[t!]
\centering
\resizebox{\linewidth}{!}{%
\begin{tabular}{l l c c}
\hline
\textbf{Model}  & \textbf{Utt. Info.} & \textbf{Target Utt.} & \textbf{Adr.}\\
\hline
\hline
GIFT \cite{Gu2023GIFTGF} &  & - & 90.18\% \\
SS-MPC$_{Encoder}$  & & 76.38\% & 89.92\% \\
\hline
SS-MPC$_{real\text{-}world}$ & -last utt. & 62.90\% & 78.40\% \\
\hline
\end{tabular}
}
\caption{Performance of predicting target-utterance of addressee in other MPC model. Unlike other models, SS-MPC Encoder does not utilize final response.}
\label{tab:structure_masked_accuracy}
\vskip -.16in
\end{table}

\subsection{Ablation Study}

\paragraph{Post-Training and Masking Probability}
Table \ref{tab:masking} shows the performance of the model as a function of the probability \(p\) of masking the target-utterance and addressee information during post-training of the encoder. From Table \ref{tab:masking}, we can see that the post-trained model with 25\% of masking probability effects impressively on final performance of generation, while 50\% of masking probability achieves almost same performance as the model only fine-tuned without post-training. The post-trained models with 75\% and 100\% of masking probability shows even lower performances than the model only fine-tuned without post-training, which can be interpreted as excessive masking degrades the model's analytical capability. This results provides the partial criteria of required masking probability when re-learning newly added token embeddings using the Masked Language Modeling (MLM) approach.

\paragraph{Addressee prediction}
The SS-MPC is trained to understand structures through structure tokens via post-training process. We hypothesize that this approach could be applied to the task of addressee prediction. To verify this, we train the model by masking only the addressee and predicting it. Table \ref{tab:accuracy} shows the comparison of the addressee prediction tasks with other MPC analysis models, in terms of Precision@1. It shows that SS-MPC achieves 89.92\% in addressee prediction, which is very close to the previous SOTA model, GIFT.

\subsection{Response Generation in Real-World MPC Scenario}
SS-MPC can generate responses even when partial structural information is missing by simply masking the absent tokens as Equation \ref{eq:omitted}.
Here, we assume the real-world MPC scenario, where the model should continuously generate the MPC. In this scenario, the target-utterance and addressee information of the response is missing. And the model has to predict the missing target-utterance and addressee while generating the following response. Our method can accumulate the predicted structure information to generate the next response continuously in this scenario. "$[IDX_{t}]$" and "$[SPK_{a}]$" are masked with structure masking tokens to construct sequence-structured input (Equation \ref{eq:final_response}).

Table \ref{tab:structure_masked} presents the performance of SS-MPC without target-utterance and addressee information for the final response, which is marked as SS-MPC$_{real\text{-}world}$ in this table. While its performance is slightly lower than SS-MPC with full structural information, it still outperforms the previous SOTA model, MADnet, in BLEU and maintains comparable scores in METEOR and ROUGE-L. Unlike existing SOTA models that require all the target-utterance and addressee information for each utterance, SS-MPC can generate the response with predicting the most appropriate target-utterance and addressee to answer itself.

In addition, Table \ref{tab:structure_masked_accuracy} demonstrates SS-MPC’s ability to predict the target-utterance and addressee of the last utterance in the real-world scenario. Although the performance is inevitably lower than in an environment where responses are present, SS-MPC still appears to maintain a reasonable performance on the target-utterance and addressee prediction in this scenario.

\section{Conclusions}

We introduce SS-MPC, a model optimized for generating responses in multi-party conversations. Unlike traditional graph-based approaches, SS-MPC employs the encoder-decoder architecture of transformer to fully leverage the pre-trained knowledge of language models. For this, we propose a novel method to encode dialogue structure sequentially within the input, allowing the model to capture the interaction flow in the dialogue without relying on explicit graph representations. SS-MPC outperforms the existing SOTA MPC response generation model and has the distinct advantage of easily application to real-world MPC scenarios depending on not requiring any additional module. 

\section*{Limitations}
The SS-MPC proposed in this paper has shown good performance compared to existing MPC response generation models, but it is limited by the fact that both training and inference are performed only on the Ubuntu IRC benchmark dataset, which makes it less generalizable. This is due to the absolute lack of MPC datasets, and it is necessary to apply the model to a wider variety of topics and conversations between different speakers to maintain generality. There is another room for further development of the model. For example, the model can be trained by initializing the initial embeddings of the added soft prompt tokens to specific values (e.g., [CLS] or [SEP] embeddings), or by initializing the embeddings to follow a specific distribution. Acquiring additional MPC datasets and further developing Multi-MPC will be part of our future work.

\bibliography{main}

\newpage

\appendix

\section{Data Statistics}
\label{sec:data_stats}
\begin{table}[h!]
    \centering
    \resizebox{\linewidth}{!}{%
    \begin{tabular}{lccc}

        \hline
        \textbf{Datasets} & \textbf{Train} & \textbf{Valid} & \textbf{Test} \\
        \hline
        \hline
        \citet{ouchi-tsuboi-2016-addressee} & 461,120 & 28,570 & 32,668 \\
        \citet{Hu2019GSNAG} & 311,725 & 5,000 & 5,000 \\
        \hline
    \end{tabular}
    }
    \caption{Statistics of the two benchmarks evaluated in this paper.}
\end{table}

\section{Case Study}
\label{sec:case_study}

\hrule
\vspace{1em}
\textbf{Case 1}
\vspace{1em}
\hrule
\noindent\textbf{}
\begin{quote}
\textbf{SPK1: } e : could not open lock file FILEPATH - open ( 13 permission denied ) e : unable to lock the administration directory ( FILEPATH are you root ?\\
\textbf{SPK2: } you need to be using sudo \\
\textbf{SPK3: } ah , ok , so it 's not just me \\
\textbf{SPK4: } you can read all about FILEPATH issues on URL \\
\textbf{SPK2: } you can edit off the us . part of the url and that should fix you up \\
\end{quote}
\hrule
\begin{quote}
\noindent\textbf{target-utterance: } e : could not open lock file FILEPATH - open ( 13 permission denied ) e : unable to lock the administration directory ( FILEPATH are you root ? \\
\noindent\textbf{answer speaker: } SPK5 \\
\noindent\textbf{answer adressee: } SPK1 \\
\\
\noindent\textbf{Generated Responses:}
\end{quote}
\begin{table}[h!]
\centering
\begin{tabular}{lp{0.67\linewidth}}
\textbf{Models} & \textbf{Response} \\ 
\hline
\hline
\textbf{Gold Label} & well , you are n't root , are n't you ? EMOJI \\
\textbf{Bart} & are you using sudo or gksudo ? \\
\textbf{HeterMPC} & please don't flood, use url to paste, do nix, do not use enter as punctuation. \\
\textbf{MADNet} & please don't flood, use URL to paste, do nt use enter as punctuation. \\
\textbf{SS-MPC} & you need to be able to open it as root \\
\end{tabular}
\label{tab:case_study1}
\end{table}

\newpage

\hrule
\vspace{1em}
\textbf{Case 2}
\vspace{1em}
\hrule
\noindent\textbf{}
\begin{quote}
\textbf{SPK1: } many iconpacks may be for the older kde3 \\
\textbf{SPK2: } i have run checkdisk from windows and i have many errors in partition \\
\textbf{SPK3: } sure , ai n't there a way to filter out 4.1 packs ? \\
\textbf{SPK1: } thats not a good sign . \\
\textbf{SPK3: } i 'm using kde to do that \\
\end{quote}
\hrule
\begin{quote}
\noindent\textbf{target-utterance: } sure , ai n't there a way to filter out 4.1 packs ? \\
\noindent\textbf{answer speaker: } SPK1 \\
\noindent\textbf{answer adressee: } SPK3 \\
\\
\noindent\textbf{Generated Responses:}
\end{quote}
\begin{table}[h!]
\centering
\begin{tabular}{lp{0.67\linewidth}}
\textbf{Models} & \textbf{Response} \\ 
\hline
\hline
\textbf{Gold Label} & i imagine it depends on FILEPATH you are looking \\
\textbf{Bart} & i 'm not sure , but i guess it 's filter by version \\
\textbf{HeterMPC} & i don't know, i'm not sure how to do it \\
\textbf{MADNet} & i don't know, i've never used kde. \\
\textbf{SS-MPC} & there is a way , but i do n't remember the kde3 way \\
\end{tabular}
\label{tab:case_study2}
\end{table}

\newpage

\hrule
\vspace{1em}
\textbf{Case 3}
\vspace{1em}
\hrule
\noindent\textbf{}
\begin{quote}
\textbf{SPK1: } shaft : not now , only with the ubuntu cd i can see the disk unpartitioned \\
\textbf{SPK2: } see when i stopped a linucx distro during install a while back everything refused to boot from cd until i sorted it out \\
\textbf{SPK1: } i gues is domething like that . how do i do that ? \\
\textbf{SPK2: } well its just a guess from previous experience , \\
\textbf{SPK2: } id try installing ubuntu , with it using all your hdd , then try installing windows over it , thats what i had to do \\
\textbf{SPK1: } you mean , i will format the disk again with windows , errasing the previous ubuntu installing ? \\
\end{quote}
\hrule
\begin{quote}
\noindent\textbf{target-utterance: } you mean , i will format the disk again with windows , errasing the previous ubuntu installing ? \\
\noindent\textbf{answer speaker: } SPK2 \\
\noindent\textbf{answer adressee: } SPK1 \\
\\
\noindent\textbf{Generated Responses:}
\end{quote}
\begin{table}[h!]
\centering
\begin{tabular}{lp{0.67\linewidth}}
\textbf{Models} & \textbf{Response} \\ 
\hline
\hline
\textbf{Gold Label} & going off my experience with what seemed the same problem yes , unless you can ask someone else that may have a technical solution for you EMOJI \\
\textbf{Bart} & i will format the disk again with windows , and erase the previous ubuntu install \\
\textbf{HeterMPC} & i think you can do that with the ubuntu cd \\
\textbf{MADNet} & i dont know, i dont use windows, so i dont know \\
\textbf{SS-MPC} & you should be able to use the entire drive again , but i dont know if it will work \\
\end{tabular}
\label{tab:case_study3}
\end{table}

\newpage

\section{Adaptation on Large Language Models}

\label{sec:llm}

\begin{table}[h!]
\centering
\resizebox{\linewidth}{!}{%
\begin{tabular}{l ccccc c}
\hline
\textbf{Models}  & \textbf{BLEU-1} & \textbf{METEOR} & \textbf{ROUGE-L} \\
\hline
\hline
Qwen2.5-3B & 33.62 & 31.89 & 33.09 \\
\textbf{SS-MPC$_{Qwen2.5-3B}$}  & \textbf{34.62} & \textbf{34.20} & \textbf{33.33} \\
\hline
\end{tabular}
}
\caption{Applied result on LLM. We use Qwen2.5-3B model for training and inference.}
\label{tab:LLM}
\end{table}

The concept of dialogue structuralization is also applicable to Large Language Model (LLM). In Table \ref{tab:LLM}, we compare the effects of dialogue structuralization in LLMs. We adopt sequence-structured input in Qwen2.5-3B. The results demonstrate that sequence-structured input significantly impacts performance. Especially, Qwen achieved an improvement of nearly 1\%p in BLEU-4 and 2.5\%p in METEOR solely by utilizing sequence-structured input. This highlights the importance of incorporating conversational structure into the input representation.

\section{License}
The data used in this paper can be found on \url{https://github.com/ryanzhumich/sirnn} (Ubuntu IRC (2016)) and \url{https://github.com/morning-dews/GSN-Dialogues} (Ubuntu IRC (2019)) 
We utilize parts of the code provided by HeterMPC\footnote{\url{https://github.com/lxchtan/HeterMPC}}, which is licensed under the Apache 2.0 License.

\end{document}